\documentclass{article}

\usepackage[preprint]{neurips_2025}

\usepackage[utf8]{inputenc} 
\usepackage[T1]{fontenc}    
\usepackage{hyperref}       
\usepackage{url}            
\usepackage{booktabs}       
\usepackage{amsfonts}       
\usepackage{nicefrac}       
\usepackage{microtype}      
\usepackage{xcolor}         

\usepackage{tabularx}
\usepackage{multirow}
\usepackage{makecell} 
\usepackage{graphicx}
\usepackage{float} 
\usepackage{longtable}

\usepackage{wrapfig}
\usepackage{needspace}

\title{ToLeaP: Rethinking Development of Tool Learning with Large Language Models}

\author{
\textbf{Haotian Chen}$^{1*}$, 
\textbf{Zijun Song}$^{2*}$, 
\textbf{Boye Niu}$^{3*}$, 
\textbf{Ke Zhang}$^{4*}$, 
\textbf{Litu Ou}$^{5*}$,
\textbf{Yaxi Lu}$^{1}$,\\
\textbf{Zhong Zhang}$^{1}$,
\textbf{Xin Cong}$^{1}$,
\textbf{Yankai Lin}$^{6\dagger}$,
\textbf{Zhiyuan Liu}$^{1\dagger}$,
\textbf{Maosong Sun}$^{1}$ \\
$^1$Department of Computer Science and Technology, Tsinghua University \\
$^2$Dalian University of Technology \quad
$^3$University of Sydney \\
$^4$Waseda University \quad
$^5$National University of Singapore \\
$^6$Gaoling School of Artificial Intelligence, Renmin University of China\\
\texttt{htchen@tsinghua.edu.cn, mrlyk423@gmail.com, liuzy@tsinghua.edu.cn}
}

\begin{document}

\maketitle

\begin{abstract}
  Tool learning, which enables large language models (LLMs) to utilize external tools effectively, has garnered increasing attention for its potential to revolutionize productivity across industries. Despite rapid development in tool learning, key challenges and opportunities remain understudied, limiting deeper insights and future advancements. In this paper, we investigate the tool learning ability of 41 prevalent LLMs by reproducing 33 benchmarks and enabling one-click evaluation for seven of them, forming a \textbf{To}ol \textbf{Lea}rning \textbf{P}latform named ToLeaP. We also collect 21 out of 33 potential training datasets to facilitate future exploration. After analyzing over 3,000 bad cases of 41 LLMs based on ToLeaP, we identify four main critical challenges: (1) benchmark limitations induce both the neglect and lack of (2) autonomous learning, (3) generalization, and (4) long-horizon task-solving capabilities of LLMs. To aid future advancements, we take a step further toward exploring potential directions, namely (1) real-world benchmark construction, (2) compatibility-aware autonomous learning, (3) rationale learning by thinking, and (4) identifying and recalling key clues. The preliminary experiments demonstrate their effectiveness, highlighting the need for further research and exploration. We make our code publicly available at \url{https://github.com/thunlp/ToLeaP}.
\end{abstract}

\renewcommand{\thefootnote}{\fnsymbol{footnote}}
\footnotetext[1]{Equal contribution.}
\footnotetext[2]{Corresponding author.}
\renewcommand{\thefootnote}{\arabic{footnote}}

\section{Introduction}
\label{sec:intro}


Tool learning with large language models (LLMs), aiming to enable LLMs to acquire human-like tool-use proficiency to tackle real-world challenges across diverse industries~\cite{qin2023tool,qu2024tool}, exhibits significant progress due to the unprecedented language understanding ability of recent LLMs~\cite{openai2023gpt4,grattafiori2024llama,deepseek-ai2025deepseekr1}. With the assistance of tools, LLMs perceive their environment, reason, take actions, and achieve goals via language-based interactions more efficiently and effectively~\cite{schick2023toolformer,wang2024voyager,liu2024toolace}.

\begin{figure}[t]
    \centering
    \includegraphics[width=1.0\linewidth]{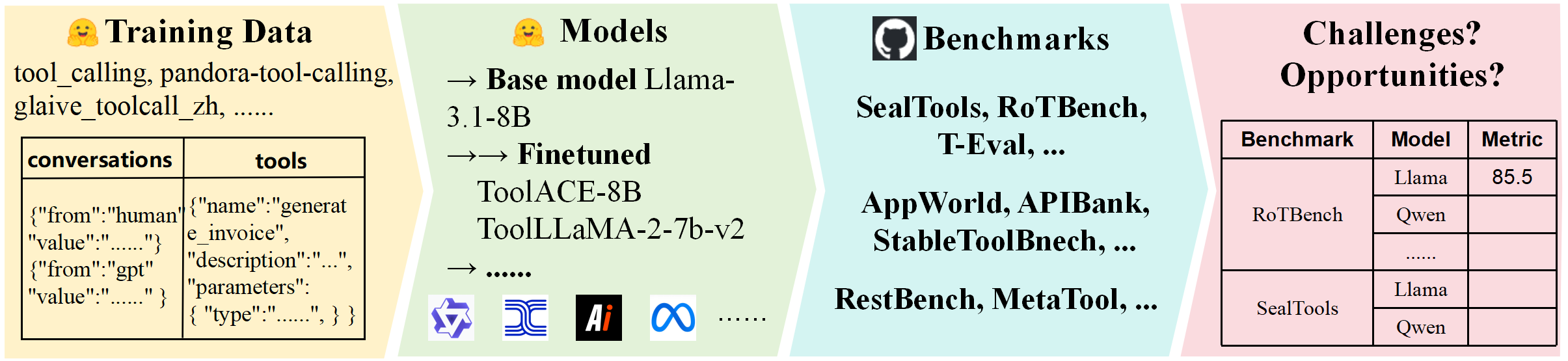} 
    \caption{A common road map of tool learning research.}
    \label{fig:roadmap}
\end{figure}

Recent work roughly divides the tool-learning capability of LLMs into four individual capabilities: task planning~\cite{wei2022chain,schick2023toolformer}, tool selection~\cite{qin2024toolllm,yuan2024craft}, tool calling~\cite{patil2024gorilla,li2024toolaugmented}, and response generation~\cite{schick2023toolformer,qin2024toolllm}. To improve these capabilities, corresponding benchmarks are proposed to provide guidance for the identification of effective methods~\cite{patil2024gorilla,shen2024taskbench,chen2024teval}, contributing to the advancements of the capabilities. For example, as shown in Figure~\ref{fig:roadmap}, based on the evaluation results from the widely recognized tool learning benchmark BFCL~\cite{patil2024gorilla}, many models, such as Qwen2.5~\cite{qwen2025qwen25} and ToolACE~\cite{liu2024toolace}, successfully demonstrate their effectiveness and infer the potential research directions. 

However, we find that existing tool-learning benchmarks struggle with providing comprehensive and faithful guidance for the evaluation and training of LLMs in the tool-learning domain.
They evaluate LLMs from the aforementioned four isolated capabilities, resulting in a fragmented assessment that fails to reflect holistic performance~\cite{qu2024tool}. For example, some efforts in improving tool-calling capability (adapting to environment-specific tool-calling language) may degrade other capabilities due to the degeneration of the common language understanding ability~\cite{qin2023tool,qin2024toolllm}. Due to the lack of standardization and the neglect of interdependencies among different capabilities, current benchmarks offer partial observations, which are often neglected and can easily lead to misleading conclusions. This obscures the core bottlenecks of tool learning, thus limiting deeper insights into recent developments and hindering accurate foresight into future directions. Therefore, establishing unified benchmarks to identify the core challenges and opportunities in the tool-learning domain is an urgent priority.

In this paper, we serve as the first effort to construct a standardized evaluation framework by enabling one-click evaluation for seven out of 33 benchmarks in the tool learning domain, attaining a higher-level perspective for holistically observing the dynamic evolution of all capabilities and revealing critical challenges. Meanwhile, we collect 21 out of 33 training datasets from current data resources and unify their data structures, facilitating our exploration of potential opportunities.

\noindent \textbf{Challenges}. Based on our constructed data platform and detailed analysis of all bad cases presented by existing models, we identify four main \textit{challenges}, where the first challenge renders the neglect of the other three challenges. (1) During model selection, \textit{benchmark limitation} causes the failure in accurately identifying the effective research attempts. (2) During the training process, LLMs struggle with \textit{autonomous learning}, failing to actively identify and select compatible data according to its current capability level, which limits the full potential and effective mining of vast amounts of valuable training data. For example, due to the absence of autonomy, increasing the training data size by 100× yields only a 5\% performance improvement over standard training. In contrast, applying simple autonomy modeling on an open-sourced subset of the data (less than 1×) achieves a 3\% improvement—nearly matching the performance of the 100× setting. (3) During the evaluation process, LLMs fail to \textit{generalize} their rich tool learning experience to learn new tools, because shortcut learning impedes them from accumulating the human-like learning experience. Specifically, ToolLLaMA-3.1-8B that successfully masters over 16,000 tools during training still exhibits an error rate exceeding 25\% when interacting with new tools from the same source. (4) In real-world applications, LLMs lack the \textit{long-horizon task-solving ability} to solve complex problems due to the insufficient key clues identification, inaccurate expectation, and limited adaptiveness of planning (strategy). For example, we manually analyze the bad cases of LLaMA-3.1-8B and GPT-4o on BFCL Multi-Turn benchmark and find that over 70\% bad cases fall under the category of missing key clues.

\noindent \textbf{Opportunities}. We take an initial step toward investigating \textit{opportunities} by exploring the potential solutions to the challenges. We find that: 
(1) while our work has alleviated the benchmark limitation to some extent, further advancements can be pursued to define a unified standard and a structured approach for \textit{real-world benchmark} design; (2) \textit{autonomously learning from compatible training data} elicits better performance when the LLM itself possesses the required baseline fundamental understanding cap, inducing up to 40\% performance improvement in accuracy of LLMs; (3) learning and following generalizable thought improves generalization capability, correcting over 50\% initial bad cases; (4) long-horizon tasks tend to be successfully solved by identifying and recalling key clues, which inverts up to 60.9\% initial bad cases of GPT-4o.

\enlargethispage{\baselineskip}
\section{Related Work}
\label{sec:related-work}
\noindent \textbf{Rethinking the Development of LLMs.}
Recently, generative LLMs such as GPT-4~\cite{openai2023gpt4}, LLaMa~\cite{grattafiori2024llama}, DeepSeek-R1~\cite{deepseek-ai2025deepseekr1} show their unprecedented success across various tasks. However, recent work reports that they are still far from achieving artificial general intelligence (AGI). For example, LLMs require over 900 tokens to answer simple questions such as ``What is 2+3?''~\cite{chen2024not}. Larger and more instructable LLMs become less reliable~\cite{zhou2024larger}. LLMs fail to improve through self-correct~\cite{huang2024large}. Furthermore, many typical issues such as hallucinations~\cite{farquhar2024detecting} and unsafe performance~\cite{zhan2024injecagent} continue to persist. In this paper, we rethink the emerging key challenges and opportunities when LLMs are required to interact with external environments and tools.

\noindent \textbf{Tool Learning with LLMs.}
Recent large language models (LLMs) show unprecedented language understanding ability~\cite{openai2023gpt4,grattafiori2024llama,deepseek-ai2025deepseekr1}, enabling them to solve complex problems via language-based tool calling~\cite{schick2023toolformer,wang2024voyager,qin2024toolllm,liu2024toolace}. Researchers improve the tool-using ability of LLMs by benchmarking and enhancing their capabilities of task planning~\cite{wei2022chain,schick2023toolformer}, tool selection~\cite{qin2024toolllm,yuan2024craft}, tool calling~\cite{patil2024gorilla,li2024toolaugmented}, and response generation~\cite{schick2023toolformer,qin2024toolllm}, respectively. 
In this paper, we collect the tool learning data resources to identify the challenges and opportunities.

\enlargethispage{\baselineskip}
\section{Data Collection}

\subsection{Preliminary}
\label{sec:preliminary}
We first clarify our categorization of benchmark reproducibility, which consists of three tiers: (1) benchmarks that meet the standards of our one-click evaluation tool learning platform ToLeaP; (2) benchmarks that are reproducible but cannot be integrated into ToLeaP; and (3) benchmarks that are currently not reproducible. We define the criteria that distinguish these categories as follows.

\noindent \textbf{Tier 1}: Benchmarks that are integrated into our unified one-click evaluation tool learning platform ToLeaP. These benchmarks meet the following conditions: (1) Although their input, inference, and evaluation procedures vary, we have successfully adapted them into a standardized paradigm within our framework, where evaluation metrics can be precisely calculated by merely giving a model; (2) The results on our revised benchmarks consistent with those in the original papers. This category includes the six benchmarks already merged into our framework, such as RoTBench, SealTools, etc. We also adopt the suggestions of reviewers, spending more cost to involve two additional benchmarks in our one-click evaluation framework.

\noindent \textbf{Tier 2}: Benchmarks that are reproducible but not compatible with the general evaluation framework. These benchmarks can reproduce the original results using their own code, but are not directly compatible with our one-click evaluation setup due to conflicts in code structure, reliance on simulation-based rather than actual tool use, environment issues, or use of non-general-purpose tools. For example, ToolBench’s original code is outdated and involves complex steps. Its dependency on online services (e.g., RapidAPI) caused tool connection failures after updates, requiring substantial revision. While we attempted to use GPT-4o for simulated tool calls, the original repository focuses only on closed-source and ToolLLaMA models, making it hard for general models (not trained on ToolBench training data) to align with ToolBench’s reasoning logic. Moreover, ToolLLaMA’s inference via HuggingFace is extremely slow (10+ minutes per output), resulting in over an hour per query on average. Our attempts to switch to vLLM inference faced major compatibility issues due to version mismatches. Hence, these benchmarks are evaluated separately.

\noindent \textbf{Tier 3}: Benchmarks with missing key data or code, making reproduction prohibitively hard. For instance, APIBench lacks evaluation code; RestBench and MetaTool are missing both evaluation code and ground-truth labels. Despite reaching out to the original authors, we did not receive any response, and thus had to exclude these benchmarks.

Due to the prohibitively expensive cost of revising the source code of some benchmarks to match their updated online environments, tools, and network settings, we do not involve them in our constructed one-click-evaluation benchmark set. We also leave the consolidation and expansion of our identified opportunities and their further validation in additional scenarios as future work.

\enlargethispage{\baselineskip}
\subsection{Data Collection Challenges}
\label{sec:data-challenges}
During the reproducing and annotating process, we encounter numerous, various, and difficult challenges. While most details are omitted, the key challenges are outlined as follows.

\noindent \textbf{Insufficient or Outdated Reproduction Information.}
The reproduction information provided in both the original paper and repository is insufficient, which has made it challenging to make the benchmark satisfy one-click-evaluation requirements despite our extensive repair and restoration. For example, the open-sourced code of MetaTool~\cite{huang2024metatool}, WTU-Eval~\cite{ning2024wtueval}, RestBench~\cite{song2023restgpt}, API-BLEND~\cite{basu2024apiblend}, API-BANK~\cite{li2023apibank}, and ToolSword~\cite{ye2024toolsword} lack either the implementation of their inference (evaluation) methods or ground truth data annotation; the online environments or APIs considered by NexusRaven~\cite{nexusflow2023nexusraven}, ToolTalk~\cite{farn2023tooltalk}, ToolBench~\cite{qin2024toolllm}, and ToolLens~\cite{qu2024completenessoriented} are updated, demanding the update of benchmark implementation; StableToolBench~\cite{guo2024stabletoolbench} only considers the evaluation of LLaMA-2 or earlier models. 

\noindent \textbf{Limited Model Coverage.}
Some benchmarks provide their own parsing rules for responses from their considered models, which renders the evaluation standards vary across different benchmarks. We have to consider the evaluation requirements and unify the inference methods, which is detailed in Appendix~\ref{appendix:inference-methods}.

\noindent \textbf{Biased Settings of Benchmarks.}
Biases in the settings of the benchmarks resulted in inaccurate evaluations. For example, in the T-Eval~\cite{chen2024teval} template setting of the LLaMa-2 model series, the end token ``[/INST]'' was mistakenly written as ``[\textbackslash INST]'', leading to incorrect formatting after template conversion. Similarly, the SealTools~\cite{wu2025sealtools} set the maximum length to 256, causing many responses to be truncated before they could generate a full response and leading to lower evaluation results. Furthermore, TaskBench~\cite{shen2024taskbench} demands a complex JSON output format, which results in frequent parsing errors and worse performance.

\subsection{Annotation and Reproducibility}
\label{sec:annotation}

\noindent \textbf{Annotation and Reproduction Guideline.}
We first investigate and present the candidate benchmarks and training datasets in two tables, which summarize their original information and serve as the indicator for quality checks of reproduction.
Each reproduction team member is required to carefully read the original paper and accurately answer all the questions raised by team members. Otherwise, the member revisits the paper and repeats the process. Failure to meet the requirements more than three times will result in replacement. All members of the final reproduction team possess research experience in the computer science (CS) domain with their own publications. Based on their experience, they further receive training on benchmark reproduction with demonstrations. Over 80\% of the team members are Postdocs or PhD students. During the reproduction of benchmarks, each member strictly follows the ``Readme.md'' from the original repositories. When debugging, any source code modifications or reproduction stop reasons must be explained, noted in the table, and approved in a unanimous team discussion. During the annotation of training datasets from Hugging Face, each team member observes ten randomly sampled data instances to confirm their relevance to the tool learning domain followed by double checks of other team members. Then, they transform the datasets into the unified ShareGPT format.

\noindent \textbf{Reproducing Process and Quality.}
During the replication process, we maintain a table of experimental results, where each row corresponds to an LLM, and each column represents the evaluation metrics defined in the original benchmark paper or repository. The results from the original paper are pre-filled into the table, separated from our reproduced results by a separator. If the error value is lower than 5\%, the cell is marked blue; between 5\% and 10\%, it is yellow; and above 10\%, it is red. A benchmark is considered successfully reproduced if at least two models are fully marked in blue. If the failure persists, we continue debugging until the cause of the error is identified. Based on the causes, we decide to either modify the benchmark or abandon it. When collecting the training data, the final unified training data format must be compatible and directly executable with pretraining frameworks defined by LLaMA-Factory, and the randomly sampled cases from each dataset are checked at least twice to ensure the data is correctly understood and consistent with the guidelines.


\enlargethispage{\baselineskip}
\vspace{-1ex} 
\section{Experiment and Analysis}

\subsection{Experimental Settings}
\label{sec:experimental-settings}
\noindent \textbf{Implementation Details.} 
We conduct experiments on eight A100 GPUs. 
To accelerate the inference speed, we adopt vLLM\footnote{\url{https://github.com/vllm-project/vllm}} with multiple processes to implement chat-completion-based inference methods and use the streaming mode of Hugging Face transformers\footnote{\url{https://huggingface.co/docs/hub/transformers}} inference methods with batch modifications to implement streaming-based methods.
All inference methods are configured with a maximum input sequence length of 4096 tokens and a temperature setting of 0.

\noindent \textbf{Evaluation Metrics.}
The benchmarks focus on evaluating various capabilities of LLMs across various tasks: (1) tool selection and use, (2) parameter identification and filling, (3) task structure and output accuracy, (4) task execution and completeness, and (5) robustness. More details are shown in Table~\ref{metric-details} in the Appendix.


\subsection{Challenges}  
\label{sec:challenges}

We conduct comprehensive experiments on our reproduced benchmarks and collected training data to identify the key challenges that impede the development of tool learning. We first evaluate the performance of common LLMs across the benchmarks as shown in Figure~\ref{fig:overall-exp-results}. More detailed results of various evaluation metrics are presented in Appendix~\ref{appendix:evaluation-results}.
Based on the experimental results and our comprehensive analysis of bad cases from the involved LLMs, we identify four major challenges that hinder the progress of tool learning with LLMs toward its ultimate goal.

\begin{figure*}[htbp]
    \centering
    \includegraphics[width=5.5in]{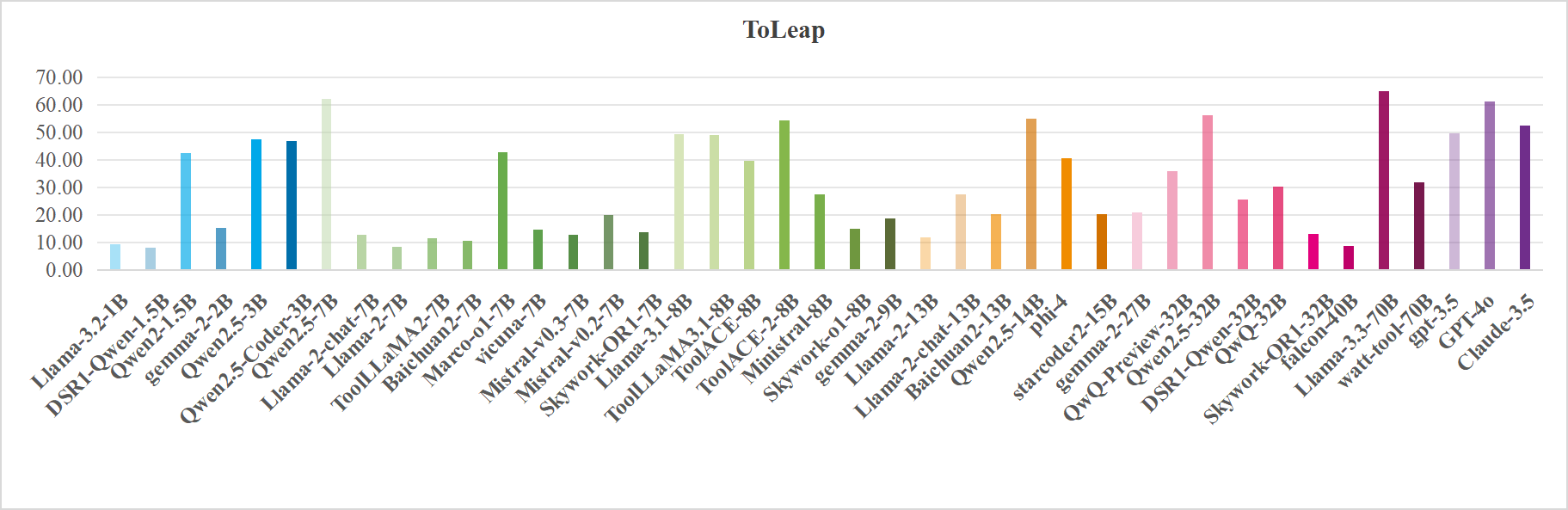} 
    \caption{Overall model performance on ToLeaP. Due to the limited space, we calculate the sample-size-weighted average performance across the total 64 metrics from all the benchmarks in ToLeaP. }
    \label{fig:overall-exp-results-toleap}
\end{figure*}





\noindent \textbf{Definition.}
In particular, \textit{real-world tool learning benchmarks} consider the real-world interaction with the dynamic environment and the evaluation of the understanding ability of environments. \textit{autonomy} of LLMs refers to their capacity to efficiently and accurately solve tasks by independently generating data through exploration and selecting data---either experience data or available training data that is appropriate to their current ability level---to learn from. \textit{Generalization} describes the ability of LLMs to apply their trained tool learning ability to learn mastering other unseen tools within the same category. \textit{Long-horizon task-solving ability} refers to identifying past key clues before taking actions, forming outcome expectations and following its guidance when taking actions, and adjusting the future plan based on new observations (e.g., achieving SOTA results in a Kaggle competition).

\noindent \textbf{Challenge 1: Limitations of Benchmark Construction.}
Different benchmarks evaluate different capabilities. After observing the data and bad cases, we identify four main \textit{limitations}.
(1) \textit{Annotation accuracy}. Based on our reproduction experience and analysis of bad cases, we observed that at least 10\% of the bad cases stem from inaccurate information within the benchmark itself, leading to an underestimation of the performance.
(2) \textit{Environmental interaction}. Many benchmarks construct simulated tools, lacking detailed execution information. These benchmarks primarily focus on evaluating whether the tool calling instructions and parameters are correct, and a few benchmarks adopt LLMs to simulate the environment without considering interacting with the real-world environment.
(3) \textit{Environmental understanding}: We found that most benchmarks do not require LLMs to understand the environment they operate in: LLMs are not required to generate executable tool using instructions, based on their environment, query, and available tools, to continuously using tools until the task is completed. Despite the attempt of BFCL-v3 that focuses on file systems (environment) and cmd commands (tools), it remains far from the more general tool environments required for broader real-world applications. 
(4) \textit{Reproduction cost}: Some benchmarks adopt online environments and APIs to model a tool using process. However, these online tools and environments are frequently updated, and different APIs sometimes require separate registration. The benchmarks model dynamic environments from a static perspective, without considering maintenance effort nor attempting to benchmark the capability of LLMs to adapt to dynamic tools and environments, thus leading to expensive reproduction costs.
More details are shown in Appendix~\ref{appendix:challenge-1}.

Due to the limitations of benchmarks, their provided feedback could be biased, rendering the research direction biased and critical issues veiled. After observing and analyzing all bad cases provided by our constructed one-click-evaluation benchmark set, we identify three primary challenges of tool learning with LLMs: autonomy, generalization, and long-horizon task solving, which prevent LLMs from achieving their goal of revolutionizing human productivity by tool learning.

%
\begin{figure}[t]
    \centering
    \includegraphics[width=1\linewidth]{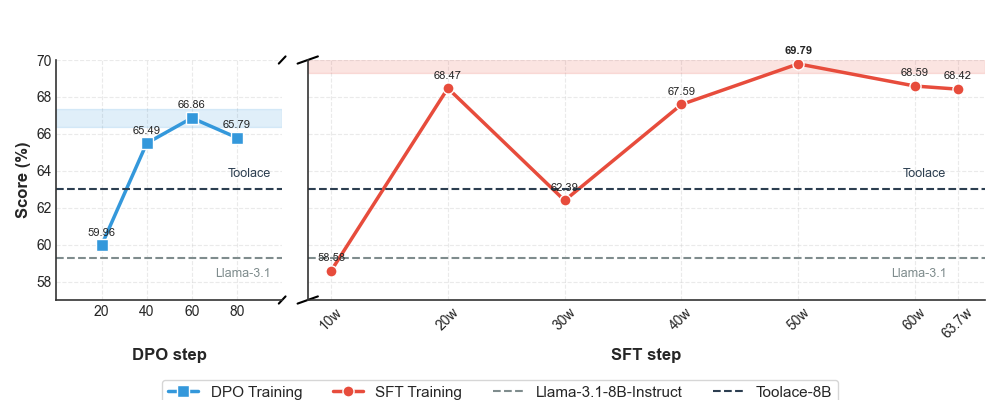} 
    \caption{Model performance on RotBench and Sealtools. The comparison of same LLM trained on selected data (ToolACE) versus full data SFT and DPO.}
    \label{fig:toolace-results}
\end{figure}

\noindent \textbf{Challenge 2: Lack of Autonomous Learning Capability.}
Since it is infeasible for humans to manually curate training data for LLMs in a dynamically evolving landscape of knowledge and tools, LLMs themselves require discerning which data is valuable, redundant, or detrimental for their learning according to their current ability level. 
However, such an autonomous learning capability is neglected by current methods, significantly impeding both the efficiency and effectiveness of tool learning with LLMs. 
\textbf{For efficiency}, we compared the performance of LLaMA-3.1-8B-instruct trained on 33 datasets against that trained solely on the ToolACE dataset, referred to as ToolACE-8B. The results reveal two key findings: (1) Simply scaling up the training data does not lead to consistent performance gains; in fact, it can cause performance degradation, suggesting the presence of data that is detrimental to model learning. (2) Even when the model is fine-tuned with 100 times (10,000\%) more training data than ToolACE dataset, the performance improvement over the ToolACE-8B baseline is limited to just 5\%. These findings indicate that training data contains a mixture of samples: some conducive to learning and others not. Without explicit differentiation or handling, such data heterogeneity significantly constrains the potential value and learning efficiency of the dataset. 
\textbf{For effectiveness}, we introduce a mechanism that simulates basic forms of proactivity in the model to examine its impact on learning effectiveness. Specifically, we use the open-sourced subset of ToolACE dataset for training, and allow the model to autonomously attempt task completion, assess its own failures before reflecting on its capability level, and retry to explore better attempts or skip the current task based on prior experience. To jointly learn both original (positive) and actively explored (negative) data, we adopted a DPO (Direct Preference Optimization) training framework. As shown in the results, the simple autonomous learning model not only outperformed the ToolACE-8B model trained on 100\% of the ToolACE dataset, but also achieved performance comparable to models trained on 10,000\% more data. To sum up, the experimental results underscore the need for models to autonomously select what to learn. More details are shown in Appendix~\ref{appendix:challenge-2}.

\noindent \textbf{Challenge 3: Lack of Generalization Capability.}
We find that most LLMs struggle to generalize their tool-learning ability to unseen tools or scenarios. For example, ToolLLaMA exhibits a significant performance gap to human tool users: Instead of behaving like a novice when given unseen tools, human tool users will develop systematic learning strategies that allow them to quickly adapt to unseen tools after their successful training on 16,000+ API tools. While ToolLLaMA, which serves as the fundamental reference in tool learning and successfully masters over 16,000 APIs from the RapidAPI platform, fails to correctly use tools that share the same source and format as its training data. Specifically, we first reproduce ToolLLaMA by updating its base model from LLaMA-2 to LLaMA-3.1. Then, we evaluate the performance of ToolLLaMA-3.1-8b.
\begin{wrapfigure}{r}{0.5\textwidth} 
    \centering
    \includegraphics[width=\linewidth]{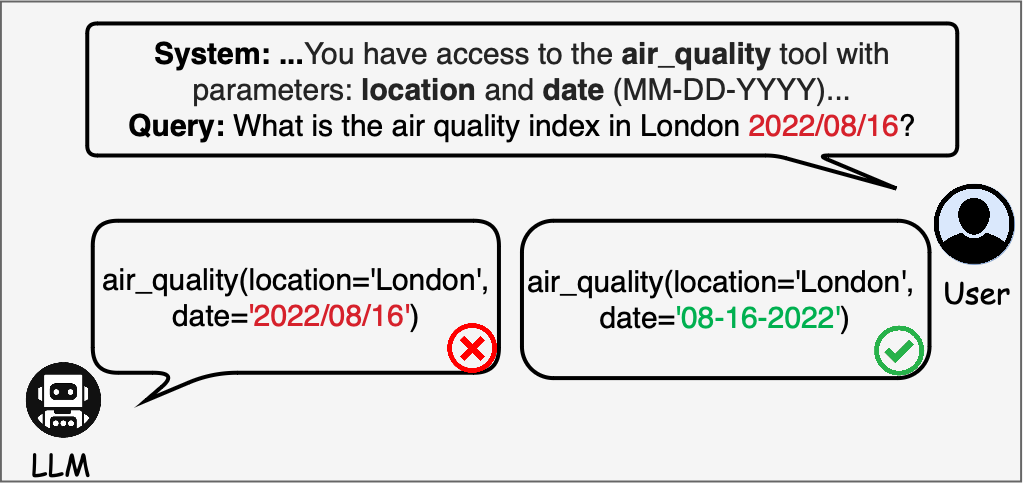}
    \caption{An illustration of shortcut learning. The LLaMa-3.1-8B-instruct-based models forward the information provided in the prompt directly as tool parameters, without performing any modifications and validity checks.}
    \label{fig:shortcut-example}
\end{wrapfigure}
On RoTBench, 86.62\% of cases failed, even though the tool call format and instructions were consistent with those used in the training set. On SealTools, the error rate was 22.91\%. Notably, when evaluated on BFCL RapidAPIs, which share the same source as the 16,000 training APIs, 25.7\% of cases (28 out of 70) still failed, despite alignment in both format and origin. More details are shown in Appendix~\ref{appendix:challenge-3}.

\noindent \textbf{Shortcut Learning.}
We further diagnose the reason why LLMs fail to acquire generalization ability during the training process. We observe the most significant error type: shortcut learning, which dominates the bad cases on BFCL. As shown in Figure~\ref{fig:shortcut-example}, LLMs fail to learn the tool learning rationale that is simply learned by humans: they do not check the format of parameters before tool calling.

\begin{figure}
    \centering
    \includegraphics[width=\linewidth, trim=0cm 1.3cm 0cm 0cm, clip]{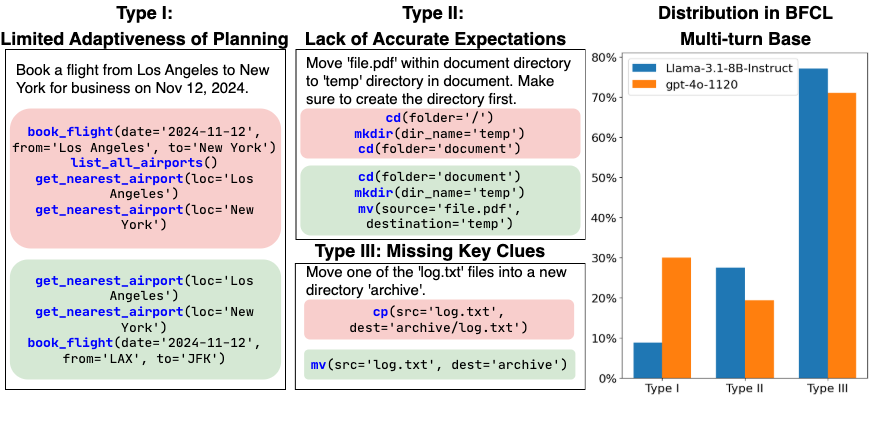}
    \caption{Model performance on 200 examples from the BFCL-v3 Multi-Turn task. We manually review all bad cases for accuracy.}
    \label{fig:bfcl_multi_turn}
\end{figure}

\noindent \textbf{Challenge 4: Lack of Long-Horizon Task-Solving Capability.}
The strong long-horizon task-solving ability makes LLMs succeed in solving a wide range of complex real-world problems and brings tremendous value. When solving long-horizon tasks, humans identify and integrate \textit{past} key clues, imagine the potential outcomes of \textit{current} intended actions to assess their relevance to the goal, and modify \textit{future} plan and expectations according to the current actual observations. In contrast, after manually annotating the bad cases on the BFCL Multi-Turn dataset on 200 examples, where Llama-3.1-8B-Instruct achieves 6.88\% accuracy and gpt-4o-1120 achieves 47.62\% accuracy as shown in Figure~\ref{fig:overall-exp-results} and Figure~\ref{fig:bfcl_multi_turn}, we demonstrate that current LLMs exhibit key limitations in solving long-horizon tasks. (1) \textit{Missing key clues}: LLMs fail to extract key clues from past experiences, leading to inconsistent actions and responses in long-term action execution. As shown in the bottom-middle part of Figure~\ref{fig:bfcl_multi_turn}, LLMs neglect the key information ``move''. (2) \textit{Lack of Accurate Expectations}: LLMs lack the ability to anticipate potential action outcomes, which not only increases the risk of unintended and irreversible environmental changes but also results in inefficient decision-making by executing actions without a goal-driven strategy. As shown in the top-middle part of Figure~\ref{fig:bfcl_multi_turn}, LLMs fail to predict the current directory after two operations and thus take an infeasible action. (3) \textit{Limited Adaptiveness of Planning}: LLMs either lack future planning or, if a plan exists, fail to adapt and implement it effectively, rendering them incapable of responding to diverse environmental feedback. As shown in the left part of Figure~\ref{fig:bfcl_multi_turn}, LLMs struggle to ground their learned general strategy into specific sequences of actions, resulting in a reversed action sequence for flight booking. To sum up, For Llama-3.1-8B-Instruct, 77.2\%, 27.5\%, and 8.8\% of the bad cases fall under the category of Type III, Type II, and Type I errors, respectively. For GPT-4o-1120, 71.1\%, 19.4\%, and 30.1\% of the bad cases fall under the category of Type III, Type II, and Type I errors, respectively.

\subsection{Opportunities}  
\label{sec:opportunities}

\noindent \textbf{Opportunity 1: Real-World Benchmark Construction.}
Based on our insights from benchmark reproduction and observations of the collected training data, we recommend a structured approach for future benchmark design. The first step is to establish a well-defined and stable environment. Given this environment, the format of tools should be specified by determining the necessary information required for their execution. Meanwhile, precise tool execution commands are formulated to ensure the tools can interact with the environment and receive accurate feedback as expected. Within the action space defined by the environment and the available tools, specific tasks are constructed to ensure a standardized and generalizable evaluation framework. A benchmark developed under this paradigm would offer several advantages: it ensures environmental stability and reproducibility, maintains high-quality and well-defined tool specifications, and enables precise interaction between tools and the environment. Based on the unified construction standard, benchmark improvements will directly contribute to real-world problem-solving without the constraint of their scope.

\begin{figure}
    \centering
    \includegraphics[width=\linewidth]{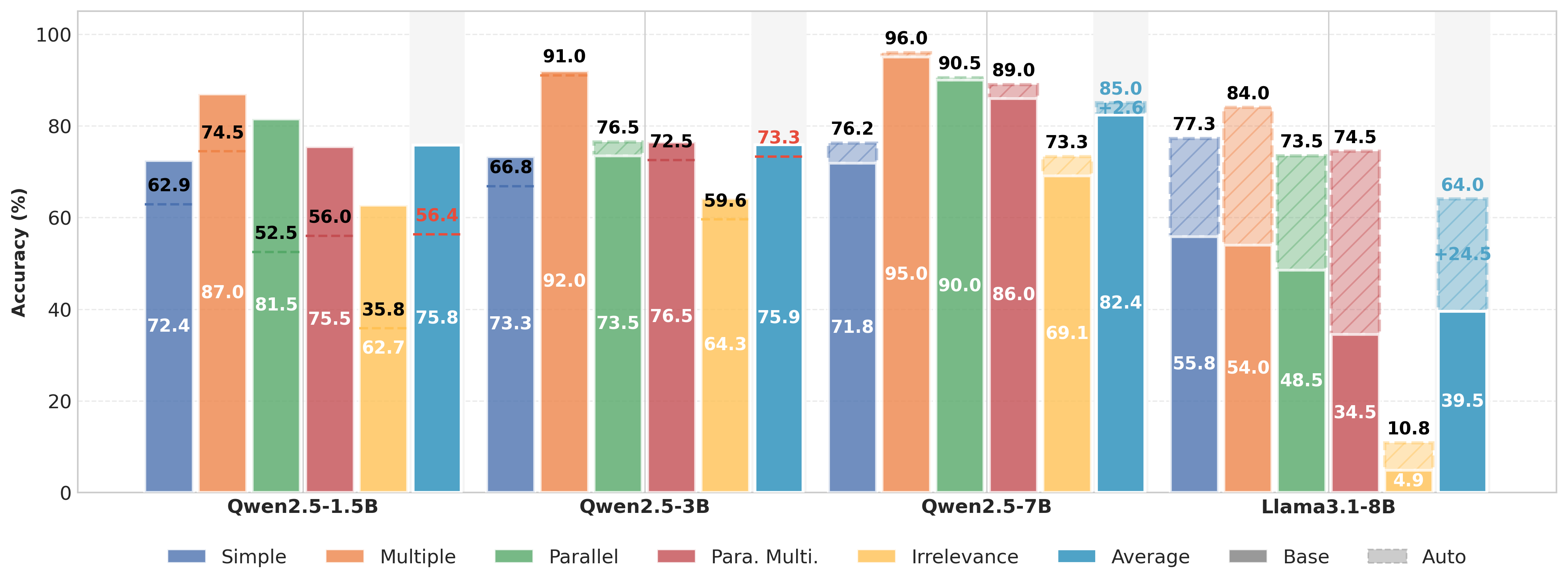}
    \caption{Results of compatibility-aware learning of different models on BFCL.}
    \label{fig:compatibility}
\end{figure}

%
\noindent \textbf{Opportunity 2: Compatibility-Aware Autonomous Learning.} 
LLMs exhibit varying capability levels, excessively complex data during post-training may induce hallucinations, whereas overly simplistic data offers limited potential for further capability improvement. Therefore, LLMs require the ability to autonomously align their learning process with their current capability level. We model autonomous behaviors in challenge 2 through self-critique and reflection, and validate the effectiveness of autonomy. In this part, we further leverage model autonomy to propose a compatibility-aware learning method that enables LLMs to assess their capabilities and select their training data through self-evaluation. Specifically, we adopt models from both Qwen and LLaMA3.1 model families as base models for compatibility-aware data selection. For each dataset in the 33 datasets, we performed interactions with randomly sampled $k$ instances and assigned the model a self-evaluation task: to assess whether it believes it is capable of producing ground-truth responses. Each instance receives a quality score from 0 to 10. Instances scoring 5 or above are considered high-quality and aligned with the model’s current capabilities. We then fine-tuned the base model using this high-quality dataset. In future iterations, the fine-tuned model replaces the base model, forming a bootstrapping loop that iteratively improves data selection and model performance. The LLM identifies and chooses data samples it deems appropriate and subsequently learn from these self-selected samples through SFT.
As shown in Fig~\ref{fig:compatibility}, we find that the performance of autonomous learning degrades on small-scale models (e.g., 1.5B-3B) while enhances on larger-scale models (e.g., 7B-8B). We attribute the performance differences to variations in the understanding and evaluation capabilities of models. Autonomous learning relies on the model’s ability to assess the quality and relevance of training data. We observe that smaller models present overly conservative data filtering, which renders many potentially useful samples discarded. In contrast, we observe that larger models have sufficient understanding and evaluation capabilities to identify and retain samples that are beneficial for further training. Therefore, LLMs will possess stronger tool learning capability through compatibility-aware autonomous learning based on the required baseline fundamental capabilities.

\noindent \textbf{Opportunity 3: Rationale Learning by Thinking.} 
Recently, o1-like LLMs demonstrate strong logical reasoning abilities. They construct a reasoning process, namely ``thinking'', that explains how they arrive at their final answer. Previous work demonstrates that ``thinking'' significantly improves the accuracy of answers. Since we find that LLMs exhibit shortcut learning without thinking, we pose the question: Can LLMs integrate the rationale of tool learning into their \textbf{thinking patterns} to enhance their generalization ability? We conduct experiments to further investigate the question. First, we observe from Figure~\ref{fig:overall-exp-results} that o1-like models, including QwQ-32B, Macro-o1, and Skywork-o1, struggle to generate precise tool-use instructions. The performance stems from either flawed thinking or a correct explanation of a thinking process that is improperly understood. We take a step further by providing LLMs with a human-annotated oracle CoT of tool learning. As illustrated in Figure~\ref{fig:thought-example}, the results indicate that correct thinking leads to improved model performance, mitigating previous errors. However, we notice that thinking raises a new concern: the model continues to perform shortcut learning from the thinking content. Our findings suggest that despite the effectiveness of thought, the improvement will be limited by the language understanding abilities of LLMs. Therefore, we appeal to future work to focus on improving the capabilities of both language understanding and thinking (with rationales), thus eliciting strong generalization ability of LLMs, namely drawing inferences from one instance. More details of CoT are presented in Appendix~\ref{appendix:3-opportunity}.



%
\noindent \textbf{Opportunity 4: Identifying and Recalling Key Clues.} 
We identified three key directions in long-horizon task solving from the perspectives of the past, present, and future: identifying past key clues, anticipating the outcomes of actions, and dynamically adjusting plans based on new feedback. Since the core foundation lies in the key clues from either anticipation or actual observation, we pose the question: Do LLMs require the explicit identification and recall of key clues, or is it sufficient to leave them in the dialogue history for the model to recognize and respond accordingly? To address the question, we conducted experiments on the BFCL-v3 benchmark based on LLaMA-3.1-8B-instruct. Specifically, we manually annotate its bad cases, identify critical information and the potential outcomes of actions, and explicitly provide the key information before the model executes its actions. For example, when the model was about to perform the next operation in the file system, we supplemented the prompt with the statement: ``You are in the /root directory.''
Our observations indicate that such interventions lead to a significant correction of bad cases: after adding these targeted hints, LLaMA-3.1-8B-Instruct corrected 52.4\% of previously failed cases. GPT-4o-1120 corrected 60.9\% of those cases. Therefore, we appeal to future work to focus on accurately and explicitly identifying and recalling those key clues from anticipation or actual observation.



\section{Conclusion}
In this paper, we integrate existing training and evaluation resources in the tool learning domain and leverage them to conduct comprehensive analyses and experiments, aiming at identifying both key bottlenecks that hinder the advancement of tool learning and promising future research directions. By analyzing all bad cases across various models and benchmarks, we identify four major challenges. Correspondingly, we conduct experiments to unravel potential research directions. We expect that the identified challenges will deepen insights, while the identified opportunities will aid the future development of tool learning research.



\bibliography{main}
\bibliographystyle{plain}

\newpage
\appendix

\begin{table}[ht]
\centering
\resizebox{\columnwidth}{!}{
\begin{tabular}{|p{0.15\textwidth}|p{0.25\textwidth}|p{0.5\textwidth}|}
\hline
\textbf{Benchmark} & \textbf{Metric} & \textbf{Metric Explanation} \\ \hline
\textbf{RoTBench} & Tool selection, Parameter identification, Content filling & Evaluates the LLM’s ability to choose the appropriate tool, recognize parameters, and populate them correctly. \\ \hline
\textbf{SealTools} & Format ACC, Tool P/R/F1, Parameter P/R/F1 & Assesses the correctness of output structure, tool selection, and accuracy in filling parameters. \\ \hline
\textbf{TaskBench} & ROUGE-1, ROUGE-2, Node F1, Edge F1, Parameter Name F1, Parameter Name \& Value F1 & Measures word/word pair matches, tool selection accuracy, dependency understanding, and correctness in parameter identification and value assignment. \\ \hline
\textbf{BFCL} & Accuracy & Assesses correctness in function invocation across various task scenarios. \\ \hline
\textbf{T-Eval} & Accuracy & Measures correctness in six task scenarios (planning, reasoning, retrieval, understanding, instruction, review). \\ \hline
\textbf{InjecAgent} & Valid rate, ASR-valid (Direct Harm), ASR-valid (S1), ASR-valid (S2) & Evaluates the model's resilience under adversarial conditions, focusing on response validity and attack success. \\ \hline
\end{tabular}
}
\caption{Overview of Metrics in Different Benchmarks}
\label{metirc-overview}
\end{table}

\begin{table}[htbp]
\centering
\caption{Details of Metrics in Different Benchmarks}
\resizebox{\columnwidth}{!}{
\begin{tabular}{p{0.15\textwidth} p{0.3\textwidth} p{0.45\textwidth}}
\hline
\multicolumn{1}{c}{\textbf{Benchmark}} & \multicolumn{1}{c}{\textbf{Metric}} & \multicolumn{1}{c}{\textbf{Metric Explanation}}                                                                                                                                                \\ \hline
\multirow{3}{*}{\textbf{RoTBench}}     & Tool selection                      & Measures whether the LLM can accurately choose the appropriate tool to address user queries.                                                                                                   \\
                                       & Parameter identification            & Assesses the LLM's ability to recognize and output the required parameters.                                                                                                                    \\
                                       & Content filling                     & Evaluates whether the LLM correctly populates the selected tool’s parameter values.                                                                                                            \\ \hline
\multirow{3}{*}{\textbf{SealTools}}    & Format ACC                          & Assesses the correctness of the model's output structure.                                                                                                                                      \\
                                       & Tool P/R/F1                         & Evaluates the model's ability to choose the correct tool.                                                                                                                                      \\
                                       & Parameter P/R/F1                    & Measures the model’s capability in accurately filling in tool parameters.                                                                                                                      \\ \hline
\multirow{6}{*}{\textbf{TaskBench}}    & ROUGE-1                             & Examines whether the model can correctly capture and generate individual word matches, reflecting the surface-level accuracy of task decomposition.                                            \\
                                       & ROUGE-2                             & Extends ROUGE-1 by evaluating adjacent word pair matches to provide a more precise assessment of task structuring.                                                                             \\
                                       & Node F1                             & Assesses the model’s accuracy in selecting the appropriate tool for each subtask.                                                                                                              \\
                                       & Edge F1                             & Evaluates the model's understanding of dependencies between tools, ensuring correct connections in complex workflows.                                                                          \\
                                       & Parameter Name F1                   & Measures whether the model correctly identifies required parameters.                                                                                                                           \\
                                       & Parameter Name \& Value F1          & Ensures that the model not only recognizes parameters but also assigns the correct values, validating the completeness and accuracy of tool configuration.                                     \\ \hline
\textbf{BFCL}                          & Accuracy                            & Focuses on assessing the model's correctness in function invocation across various task scenarios, including simple, multiple, parallel, multiple-parallel, irrelevance, and multi-turn tasks. \\ \hline
\textbf{T-Eval}                        & Accuracy                            & Measures the model's correctness across six task scenarios: planning, reasoning, retrieval, understanding, instruction, and review. Includes evaluations in JSON and string formats.           \\ \hline
\multirow{5}{*}{\textbf{InjecAgent}}   & Valid rate                          & Measures the proportion of responses that are both non-empty and correctly formatted under attack scenarios.                                                                                   \\
                                       & ASR-valid                           & Quantifies the proportion of successful attacks within valid responses, offering a finer-grained evaluation of model vulnerability.                                                            \\
                                       & ASR-valid (Direct Harm)             & Evaluates the model’s susceptibility to direct harm attacks, where it executes malicious tool-based instructions.                                                                              \\
                                       & ASR-valid (S1)                      & Assesses the success rate of attacks under a specific condition (S1).                                                                                                                          \\
                                       & ASR-valid (S2)                      & Assesses the success rate of attacks under another specific condition (S2).                                                                                                                    \\ \hline
\end{tabular}
}
\label{metric-details}
\end{table}

\section{Details of Data Collection}
\noindent \textbf{Motivation.}
Our motivation is to deepen insights into current work and thus facilitate foresight into future directions in the tool learning domain. To this end, we attempt to reproduce the existing benchmarks and collect the training and test data to gather as comprehensive information (e.g., overall performance and bad cases) as possible to analyze the challenges and opportunities of tool learning. Specifically, we first attempt to reproduce all the existing tool learning benchmarks to construct the one-click-evaluation benchmark set. The final version of the set comprises 7 out of 33 benchmarks and possesses the functionality that takes an LLM as input and outputs the values of all evaluation metrics proposed by the benchmarks. Meanwhile, we manually collect the existing training datasets in the tool learning domain to facilitate the development of future tuning-based methods.

\section{Details of Related Work}
Abundant previous work rethinks the development of LLMs.
Traditionally, researchers identified the limitations of large language models (LLMs) by investigating their decision-making rules~\cite{mudrakarta2018did,geirhos2020shortcut}, generalization~\cite{fu2020rethinking,chen2024oodreb}, robustness~\cite{wang2022identifying,chen2023did}, and task performance~\cite{cui2022stable,chen2024rethinking}. The limitations stem from the lack of (1) effective data governance, (2) causality-driven LLM decision rules, and (3) causality-aware benchmarks, which renders LLM learning spurious correlations. 

As to tool learning with LLMs,
related methods can be roughly divided into two categories: tuning-free and tuning-based methods. Tuning-free methods typically develop agent workflows such as CoT~\cite{wei2022chain}, self-refine~\cite{wang2024voyager,yuan2024craft}, and multi-agent collaboration~\cite{shi2024learning} based on test-time computation, while tuning-based methods usually perform supervised fine-tuning~\cite{schick2023toolformer,qin2024toolllm,liu2024toolace} based on their curated training data.

\section{Details of Evaluation Metrics}
\label{appendix:evaluation-metrics}

As shown in Table~\ref{metirc-overview}. We present the various evaluation metrics adopted by the involved benchmarks in the tool learning domain. Meanwhile, we elaborate on the exact meaning of each evaluation metric in Table~\ref{metric-details}. Specifically,
In RoTBench, \textbf{Tool selection} measures whether the LLM can accurately choose the appropriate tool to address user queries, \textbf{parameter identification} assesses its ability to recognize and output the required parameters, and \textbf{content filling} evaluates whether the model correctly populates the selected tool’s parameter values.  
In SealTools, \textbf{Format ACC} assesses the correctness of the model's output structure, while \textbf{Tool P/R/F1} evaluates the model's ability to choose the correct tool. \textbf{Parameter P/R/F1}, on the other hand, measures the model’s capability in accurately filling in tool parameters.  
In TaskBench, \textbf{ROUGE-1} examines whether the model can correctly capture and generate individual word matches, reflecting the surface-level accuracy of task decomposition, while \textbf{ROUGE-2} extends this by evaluating adjacent word pair matches to provide a more precise assessment of task structuring. \textbf{Node F1} assesses the model’s accuracy in selecting the appropriate tool for each subtask, and \textbf{Edge F1} evaluates its understanding of dependencies between tools, ensuring correct connections in complex workflows. \textbf{Parameter Name F1} measures whether the model correctly identifies required parameters, whereas \textbf{Parameter Name \& Value F1} further ensures that, in addition to recognizing parameters, the model assigns the correct values, thereby validating the completeness and accuracy of tool configuration.  
The evaluation framework for BFCL focuses on \textbf{accuracy} as the primary metric, assessing the model’s correctness in function invocation across various task scenarios, including simple, multiple, parallel, multiple-parallel, irrelevance, and multi-turn tasks.
T-Eval uses accuracy as the primary evaluation metric, measuring the model’s \textbf{correctness} across six task scenarios: planning, reasoning, retrieval, understanding, instruction, and review. Each task, except review, is assessed in two formats: JSON, which requires structured outputs containing tool names and parameters, and string (str), which allows more flexible textual responses.
InjecAgent primarily assesses the model’s resilience under adversarial conditions, focusing on the validity of responses and the success rate of attacks. \textbf{Valid rate} measures the proportion of responses that are both non-empty and correctly formatted under attack scenarios. Attack success rate (ASR-valid) specifically quantifies the proportion of successful attacks within valid responses, offering a finer-grained evaluation of model vulnerability. \textbf{ASR-valid} is further categorized into specific attack types: \textbf{ASR-valid (Direct Harm)} evaluates the model’s susceptibility to direct harm attacks, where it executes malicious tool-based instructions; \textbf{ASR-valid (S1)} and \textbf{ASR-valid (S2)} respectively assess the success rates of the first and second stages of data-stealing attacks, corresponding to data extraction and data transmission. \textbf{ASR-valid (Data Stealing)} aggregates the results of S1 and S2 to provide a comprehensive measure of vulnerability to data theft, while \textbf{ASR-valid (Total)} encapsulates the overall attack success rate across all tested adversarial scenarios.

\begin{table*}[htbp]
    \centering
    \caption{Overall experimental results on RoTBench.}
    \resizebox{\textwidth}{!}{
\begin{tabular}{lccccccccccccccc}
    \hline
    \multicolumn{1}{c}{\multirow{3}{*}{model name}} & \multicolumn{15}{c}{RoTBench}                                                                                                                        \\
    \multicolumn{1}{c}{}                            & \multicolumn{3}{c}{clean} & \multicolumn{3}{c}{heavy} & \multicolumn{3}{c}{medium} & \multicolumn{3}{c}{slight} & \multicolumn{3}{c}{union} \\
    \multicolumn{1}{c}{}                            & TS      & PI     & CF     & TS      & PI     & CF     & TS      & PI      & CF     & TS      & PI      & CF     & TS      & PI     & CF     \\ \hline
    \multicolumn{16}{l}{6B- Open-source LLMs}                                                                                                                                                     \\ \hline
    Llama-3.2-1B                                    & 1.90    & 0.00   & 0.00   & 0.48    & 0.00   & 0.00   & 1.43    & 0.00    & 0.00   & 2.38    & 1.43    & 1.43   & 0.95    & 0.00   & 0.00   \\
    DeepSeek-R1-Distill-Qwen-1.5B                   & 2.86    & 0.95   & 0.00   & 2.86    & 0.48   & 0.00   & 1.90    & 0.48    & 0.48   & 0.95    & 0.00    & 0.00   & 0.00    & 0.00   & 0.00   \\
    Qwen2-1.5B-Instruct                             & 58.10   & 30.48  & 20.95  & 44.29   & 19.05  & 10.95  & 50.48   & 23.81   & 12.86  & 48.10   & 23.33   & 14.76  & 41.90   & 15.24  & 5.71   \\
    gemma-2-2b-it                                   & 32.86   & 11.43  & 6.19   & 48.10   & 17.14  & 10.48  & 43.33   & 16.67   & 11.90  & 32.38   & 10.48   & 5.71   & 32.38   & 10.48  & 5.71   \\
\hline
    \end{tabular}
    }
\label{exp:overall-results-table-RoTBench}
\end{table*}

\begin{table*}[htbp]
    \centering
    \caption{Overall experimental results on SealTools.}
    \resizebox{\textwidth}{!}{
        \begin{tabular}{lccccccccccccccccccccc}
        \hline
        \multicolumn{1}{c}{\multirow{3}{*}{model name}} & \multicolumn{21}{c}{SealTools}                                                                                                                                                                                     \\
        \multicolumn{1}{c}{}                            & \multicolumn{7}{c}{dev.json}                                         & \multicolumn{7}{c}{test in domain.json}                              & \multicolumn{7}{c}{test out domain.json}                             \\
        \multicolumn{1}{c}{}                            & AMOUNT & P\_api & R\_api & F1\_api & P\_param & R\_param & F1\_param & AMOUNT & P\_api & R\_api & F1\_api & P\_param & R\_param & F1\_param & AMOUNT & P\_api & R\_api & F1\_api & P\_param & R\_param & F1\_param \\ \hline
        6B- Open-source LLMs                            & \multicolumn{21}{c}{}                                                                                                                                                                                              \\ \hline
        Llama-3.2-1B                                    & 13.86  & 62.35  & 11.75  & 19.78   & 36.78    & 8.50     & 13.81     & 13.43  & 61.77  & 11.25  & 19.04   & 40.73    & 8.70     & 14.33     & 7.95   & 65.92  & 6.09   & 11.15   & 45.99    & 5.39     & 9.65      \\
        DeepSeek-R1-Distill-Qwen-1.5B                   & 75.10  & 69.10  & 65.69  & 67.35   & 36.28    & 35.18    & 35.72     & 74.57  & 67.03  & 64.67  & 65.83   & 34.35    & 34.57    & 34.46     & 76.61  & 69.30  & 60.71  & 64.72   & 36.95    & 34.30    & 35.57     \\
        Qwen2-1.5B-Instruct                             & 95.29  & 89.13  & 82.71  & 85.80   & 69.66    & 72.93    & 71.26     & 94.57  & 88.47  & 81.67  & 84.94   & 69.94    & 72.19    & 71.04     & 93.88  & 89.14  & 78.37  & 83.41   & 70.82    & 72.15    & 71.48     \\
\hline
        \end{tabular}
    }
\label{exp:overall-results-table-SealTools}
\end{table*}

\begin{table*}[htbp]
    \centering
    \caption{Overall experimental results on TaskBench.}
    \resizebox{\textwidth}{!}{
        \begin{tabular}{lcccccccccccccccccc}
\hline
\multicolumn{1}{c}{\multirow{3}{*}{model name}} & \multicolumn{18}{c}{TaskBench}                                                                                                                                              \\
\multicolumn{1}{c}{}                            & \multicolumn{6}{c}{dailylifeapis}                       & \multicolumn{6}{c}{huggingface}                         & \multicolumn{6}{c}{multimedia}                          \\
\multicolumn{1}{c}{}                            & rouge 1 & rouge 2 & name\_f1 & t\_f1 & v\_f1 & link\_f1 & rouge 1 & rouge 2 & name\_f1 & t\_f1 & v\_f1 & link\_f1 & rouge 1 & rouge 2 & name\_f1 & t\_f1 & v\_f1 & link\_f1 \\ \hline
6B- Open-source LLMs                            & \multicolumn{18}{c}{}                                                                                                                                                       \\ \hline
Llama-3.2-1B                                    & 0.01    & 0.00    & 0.02     & 0.02  & 0.02  & 0.00     & 0.06    & 0.03    & 0.00     & 0.00  & 0.00  & 0.00     & 0.08    & 0.03    & 0.05     & 0.04  & 0.01  & 0.00     \\
DeepSeek-R1-Distill-Qwen-1.5B                   & 0.26    & 0.13    & 0.10     & 0.03  & 0.03  & 0.00     & 0.01    & 0.00    & 0.00     & 0.00  & 0.00  & 0.00     & 0.00    & 0.00    & 0.00     & 0.00  & 0.00  & 0.00     \\
Qwen2-1.5B-Instruct                             & 29.79   & 12.89   & 16.17    & 11.88 & 8.18  & 1.54     & 34.06   & 12.55   & 20.02    & 16.02 & 7.76  & 1.89     & 30.04   & 11.39   & 24.84    & 21.29 & 10.49 & 3.11     \\
gemma-2-2b-it                                   & 42.69   & 20.40   & 37.38    & 35.19 & 24.66 & 2.01     & 35.18   & 15.43   & 43.10    & 37.98 & 21.33 & 6.06     & 34.55   & 15.53   & 43.83    & 38.55 & 20.97 & 8.94     \\
\hline
\end{tabular}
    }
\label{exp:overall-results-table-TaskBench}
\end{table*}

\begin{table*}[htbp]
    \centering
    \caption{Overall experimental results on BFCL.}
    \resizebox{\textwidth}{!}{
        %
          & 94.50    & 92.00    & 85.00             & 75.00       & 67.53        & 76.36  & 74.93    & 62.50    & 66.67             & 55.78       & 88.89     & 6.12         & 8.00                    & 9.00              & 1.50                       & 6.00                     \\
Llama-2-7b-chat-hf                              & 0.00        & 0.00   & 0.00                                                                                  & 0.00     & 0.00     & 0.00              & 100.00      & 0.00         & 0.00   & 0.00     & 0.00     & 0.00              & 100.00      & 0.00      & 0.00         & 0.00                    & 0.00              & 0.00                       & 0.00                     \\
Llama-2-7b-hf                                   & 0.00        & 0.00   & 0.00                                                                                  & 0.00     & 0.00     & 0.00              & 100.00      & 0.00         & 0.00   & 0.00     & 0.00     & 0.00              & 100.00      & 0.00      & 0.00         & 0.00                    & 0.00              & 0.00                       & 0.00                     \\
ToolLLaMA-2-7b-v2                               & 0.00        & 0.00   & 0.00                                                                                  & 0.00     & 0.00     & 0.00              & 100.00      & 0.00         & 0.00   & 0.00     & 0.00     & 0.00              & 100.00      & 0.00      & 0.00         & 0.00                    & 0.00              & 0.00                       & 0.00                     \\
Baichuan2-7B-Chat                               & N/A         & N/A    & N/A                                                                                   & N/A      & N/A      & N/A               & N/A         & N/A          & N/A    & N/A      & N/A      & N/A               & N/A         & N/A       & N/A          & N/A                     & N/A               & N/A                        & N/A                      \\
Marco-o1                                        & 0.00        & 0.00   & 0.00                                                                                  & 0.00     & 0.00     & 0.00              & 100.00      & 0.00         & 0.00   & 0.00     & 0.00     & 0.00              & 100.00      & 0.00      & 0.00         & 0.00                    & 0.00              & 0.00                       & 0.00                     \\
vicuna-7b-v1.5                                  & N/A         & N/A    & N/A                                                                                   & N/A      & N/A      & N/A               & N/A         & N/A          & N/A    & N/A      & N/A      & N/A               & N/A         & N/A       & N/A          & N/A                     & N/A               & N/A                        & N/A                      \\
Mistral-7B-Instruct-v0.3                        & 0.00        & 0.00   & 0.00                                                                                  & 0.00     & 0.00     & 0.00              & 100.00      & 0.00         & 0.00   & 0.00     & 0.00     & 0.00              & 100.00      & 0.00      & 0.00         & 0.00                    & 0.00              & 0.00                       & 0.00                     \\
Mistral-7B-Instruct-v0.2                        & 0.00        & 0.00   & 0.00                                                                                  & 0.00     & 0.00     & 0.00              & 100.00      & 0.00         & 0.00   & 0.00     & 0.00     & 0.00              & 100.00      & 0.00      & 0.00         & 0.00                    & 0.00              & 0.00                       & 0.00                     \\
Skywork-OR1-7B-Preview                          & 0.00        & 0.00   & 0.00                                                                                  & 0.00     & 0.00     & 0.00              & 100.00      & 0.00         & 0.00   & 0.00     & 0.00     & 0.00              & 100.00      & 0.00      & 0.00         & 0.00                    & 0.00              & 0.00                       & 0.00                     \\
Llama-3.1-8B-Instruct                           & 84.75       & 73.00  & \begin{tabular}[c]{@{}c@{}}python 94.00\\ Java 59.00\\ JS 66.00\end{tabular}          & 93.00    & 88.00    & 85.00             & 54.17       & 61.13        & 74.42  & 73.41    & 50.00    & 50.00             & 42.74       & 77.78     & 6.88         & 6.50                    & 10.00             & 2.50                       & 8.50                     \\
ToolLLaMA-3.1-8B                                & 84.98       & 72.92  & \begin{tabular}[c]{@{}c@{}}python 93.75\\ java 59.00\\ javascripts 66.00\end{tabular} & 94.00    & 87.50    & 85.50             & 54.58       & 60.86        & 74.03  & 72.74    & 56.25    & 50.00             & 42.86       & 77.78     & 7.25         & 9.50                    & 10.50             & 8.50                       & 0.50                     \\
ToolACE-8B                                      & 87.54       & 76.67  & \begin{tabular}[c]{@{}c@{}}python 91.00\\ java 65.00\\ JS 74.00\end{tabular}          & 93.50    & 90.50    & 89.50             & 93.33       & 78.63        & 73.26  & 76.54    & 81.25    & 70.83             & 82.77       & 83.33     & 8.12         & 11.50                   & 9.00              & 6.50                       & 5.50                     \\
ToolACE-2-Llama-3.1-8B                          & 87.48       & 75.42  & \begin{tabular}[c]{@{}c@{}}python 88.25\\ java 64.00\\ javascripts 74.00\end{tabular} & 91.00    & 93.00    & 90.50             & 95.42       & 79.92        & 71.71  & 79.11    & 81.25    & 54.17             & 84.13       & 72.22     & 37.12        & 27.50                   & 49.50             & 41.00                      & 30.50                    \\
Ministral-8B-Instruct-2410                      & 0.00        & 0.00   & 0.00                                                                                  & 0.00     & 0.00     & 0.00              & 100.00      & 39.14        & 0.00   & 0.09     & 0.00     & 4.17              & 99.66       & 0.00      & 0.00         & 0.00                    & 0.00              & 0.00                       & 0.00                     \\
Skywork-o1-Open-Llama-3.1-8B                    & 0.00        & 0.00   & 0.00                                                                                  & 0.00     & 0.00     & 0.00              & 100.00      & 0.00         & 0.00   & 0.00     & 0.00     & 0.00              & 100.00      & 0.00      & 0.00         & 0.00                    & 0.00              & 0.00                       & 0.00                     \\
gemma-2-9b-it                                   & 0.00        & 0.00   & 0.00                                                                                  & 0.00     & 0.00     & 0.00              & 100.00      & 0.00         & 0.00   & 0.00     & 0.00     & 0.00              & 100.00      & 0.00      & 0.00         & 0.00                    & 0.00              & 0.00                       & 0.00                     \\ \hline
10B+ Open-source LLMs                           & \multicolumn{19}{c}{}                                                                                                                                                                                                                                                                                                                                                                 \\ \hline
Llama-2-13b-hf                                  & 0.00        & 0.00   & 0.00                                                                                  & 0.00     & 0.00     & 0.00              & 100.00      & 0.00         & 0.00   & 0.00     & 0.00     & 0.00              & 100.00      & 0.00      & 0.00         & 0.00                    & 0.00              & 0.00                       & 0.00                     \\
Llama-2-13b-chat-hf                             & 0.00        & 0.00   & 0.00                                                                                  & 0.00     & 0.00     & 0.00              & 100.00      & 0.00         & 0.00   & 0.00     & 0.00     & 0.00              & 100.00      & 0.00      & 0.00         & 0.00                    & 0.00              & 0.00                       & 0.00                     \\
Baichuan2-13B-Chat                              & N/A         & N/A    & N/A                                                                                   & N/A      & N/A      & N/A               & N/A         & N/A          & N/A    & N/A      & N/A      & N/A               & N/A         & N/A       & N/A          & N/A                     & N/A               & N/A                        & N/A                      \\
Qwen2.5-14B-Instruct                            & 86.29       & 73.67  & \begin{tabular}[c]{@{}c@{}}python 96.00\\ java 57.00\\ javascrips 68.00\end{tabular}  & 93.50    & 92.00    & 86.00             & 81.67       & 74.23        & 74.81  & 75.97    & 62.50    & 66.67             & 72.34       & 77.78     & 11.12        & 13.50                   & 18.00             & 2.50                       & 10.50                    \\
phi-4                                           & 65.77       & 59.58  & \begin{tabular}[c]{@{}c@{}}python 48.75\\ java 58.00\\ javascrips 72.00\end{tabular}  & 59.50    & 75.00    & 69.00             & 81.25       & 51.98        & 41.86  & 49.57    & 25.00    & 45.83             & 58.39       & 55.56     & 0.38         & 0.50                    & 0.50              & 0.50                       & 0.00                     \\
starcoder2-15b-instruct-v0.1                    & N/A         & N/A    & N/A                                                                                   & N/A      & N/A      & N/A               & N/A         & N/A          & N/A    & N/A      & N/A      & N/A               & N/A         & N/A       & N/A          & N/A                     & N/A               & N/A                        & N/A                      \\ \hline
20B+ Open-source LLMs                           & \multicolumn{19}{c}{}                                                                                                                                                                                                                                                                                                                                                                 \\ \hline
gemma-2-27b-it                                  & 0.00        & 0.00   & 0.00                                                                                  & 0.00     & 0.00     & 0.00              & 100.00      & 0.00         & 0.00   & 0.00     & 0.00     & 0.00              & 100.00      & 0.00      & 0.00         & 0.00                    & 0.00              & 0.00                       & 0.00                     \\
QwQ-32B-Preview                                 & 1.38        & 0.00   & \begin{tabular}[c]{@{}c@{}}python 3.00\\ java 0.00\\ javascrips 0.00\end{tabular}     & 4.00     & 0.50     & 0.00              & 100.00      & 40.92        & 7.75   & 2.85     & 0.00     & 0.00              & 98.75       & 0.00      & 0.00         & 0.00                    & 0.00              & 0.00                       & 0.00                     \\
Qwen2.5-32B-Instruct                            & 85.54       & 69.17  & \begin{tabular}[c]{@{}c@{}}python 96.50\\ java 51.00\\ javascripts 60.00\end{tabular} & 95.00    & 90.00    & 88.00             & 79.58       & 73.92        & 81.78  & 78.25    & 62.50    & 58.33             & 66.55       & 100.00    & 16.75        & 20.50                   & 25.50             & 6.50                       & 14.50                    \\
DeepSeek-R1-Distill-Qwen-32B                    & 0.00        & 0.00   & 0.00                                                                                  & 0.00     & 0.00     & 0.00              & 100.00      & 0.00         & 0.00   & 0.00     & 0.00     & 0.00              & 100.00      & 0.00      & 0.00         & 0.00                    & 0.00              & 0.00                       & 0.00                     \\
QwQ-32b                                         & 0.00        & 0.00   & 0.00                                                                                  & 0.00     & 0.00     & 0.00              & 100.00      & 0.00         & 0.00   & 0.00     & 0.00     & 0.00              & 100.00      & 0.00      & 0.00         & 0.00                    & 0.00              & 0.00                       & 0.00                     \\
Skywork-OR1-32B-Preview                         & 0.00        & 0.00   & 0.00                                                                                  & 0.00     & 0.00     & 0.00              & 100.00      & 0.00         & 0.00   & 0.00     & 0.00     & 0.00              & 100.00      & 0.00      & 0.00         & 0.00                    & 0.00              & 0.00                       & 0.00                     \\
falcon-40b-instruct                             & N/A         & N/A    & N/A                                                                                   & N/A      & N/A      & N/A               & N/A         & N/A          & N/A    & N/A      & N/A      & N/A               & N/A         & N/A       & N/A          & N/A                     & N/A               & N/A                        & N/A                      \\
Llama-3.3-70B-Instruct                          & 85.21       & 74.83  & 
\begin{tabular}[c]{@{}c@{}}python 94.50\\ java 58.00\\ javascripts 72.00\end{tabular} & 95.00    & 84.50    & 86.50             & 57.92       & 62.91        & 82.17  & 77.02    & 93.75    & 66.67             & 39.00       & 100.00    & 13.25        & 17.50                   & 15.00             & 11.00                      & 9.50                     \\
watt-tool-70B                                   & 84.92       & 78.67  & \begin{tabular}[c]{@{}c@{}}python 98.00\\ java 64.00\\ javascripts 74.00\end{tabular} & 94.50    & 86.00    & 80.50             & 84.17       & 77.65        & 86.05  & 83.48    & 81.25    & 66.67             & 68.14       & 94.44     & 55.88        & 54.50                   & 67.50             & 55.50                      & 46.00                    \\ \hline
Proprietary LLMs                                & \multicolumn{19}{c}{}                                                                                                                                                                                                                                                                                                                                                                 \\ \hline
gpt-3.5-turbo                                   & 50.92       & 57.67  & \begin{tabular}[c]{@{}c@{}}python 61.00\\ java 54.00\\ javascripts 58.00\end{tabular} & 69.50    & 39.00    & 37.50             & 85.83       & 57.26        & 38.37  & 43.21    & 37.50    & 29.17             & 81.07       & 38.89     & 3.00         & 4.50                    & 3.00              & 2.00                       & 2.50                     \\
GPT-4o                                          & 88.10       & 79.42  & N/A                                                                                   & 95.50    & 94.00    & 83.50             & 83.76       & N/A          & N/A    & N/A      & N/A      & N/A               & N/A         & N/A       & 47.62        & N/A                     & N/A               & N/A                        & N/A                      \\
Claude-3.5-sonnet                               & 72.48       & 81.42  & N/A                                                                                   & 92.00    & 70.50    & 46.00             & 64.40       & N/A          & N/A    & N/A      & N/A      & N/A               & N/A         & N/A       & 7.50         & N/A                     & N/A               & N/A                        & N/A                      \\ \hline
\end{tabular}
    }
\label{exp:overall-results-table-BFCL}
\end{table*}

\begin{table*}[htbp]
    \centering
    \caption{Overall experimental results on T-Eval.}
    \resizebox{\textwidth}{!}{
        \begin{tabular}{lccccccccccc}
\hline
\multicolumn{1}{c}{\multirow{3}{*}{model name}} & \multicolumn{11}{c}{Teval}                                                                                                                                                                                                                                                                                                                               \\
\multicolumn{1}{c}{}                            & \multirow{2}{*}{Instruct str} & \multirow{2}{*}{Instruct json} & \multirow{2}{*}{Plan str} & \multirow{2}{*}{Plan json} & \multirow{2}{*}{Reason str} & \multirow{2}{*}{Reason json} & \multirow{2}{*}{Retrieve str} & \multirow{2}{*}{Retrieve json} & \multirow{2}{*}{Understand str} & \multirow{2}{*}{Understand json} & \multirow{2}{*}{Review str} \\
\multicolumn{1}{c}{}                            &                               &                                &                           &                            &                             &                              &                               &                                &                                 &                                  &                             \\ \hline
6B- Open-source LLMs                            & \multicolumn{11}{c}{}                                                                                                                                                                                                                                                                                                                                    \\
\hline
\end{tabular}
    }
\label{exp:overall-results-table-Teval}
\end{table*}

\begin{table*}[htbp]
    \centering
    \caption{Overall experimental results on InjecAgent.}
    \resizebox{\textwidth}{!}{
        \begin{tabular}{lccccccccccc}
\hline
\multicolumn{1}{c}{\multirow{3}{*}{model name}} & \multicolumn{11}{c}{InjecAgent}                                                                                                                                                                                          \\
\multicolumn{1}{c}{}                            & \multicolumn{11}{c}{base}                                                                                                                                                                                                \\
\multicolumn{1}{c}{}                            & Valid Rate & ASR-valid (Direct Harm) & ASR-valid (S1) & ASR-valid (S2) & ASR-valid (Data Stealing) & ASR-valid (Total) & ASR-all (Direct Harm) & ASR-all (S1) & ASR-all (S2) & ASR-all (Data Stealing) & ASR-all (Total) \\ \hline
6B- Open-source LLMs                            & \multicolumn{11}{c}{}                                                                                                                                                                                                    \\ \hline
Llama-3.2-1B                                    & 27.10      & 60.40                   & 55.20          & 84.10          & 41.70                     & 51.30             & 15.90                 & 17.40        & 67.10        & 11.70                   & 13.90           \\
DeepSeek-R1-Distill-Qwen-1.5B                   & N/A        & N/A                     & N/A            & N/A            & N/A                       & N/A               & N/A                   & N/A          & N/A          & N/A                     & N/A             \\
Qwen2-1.5B-Instruct                             & 16.20      & 38.90                   & 63.30          & 90.60          & 47.50                     & 42.00             & 8.20                  & 9.40         & 58.00        & 5.40                    & 6.80            \\
\hline
\end{tabular}
    }
\label{exp:overall-results-table-InjectAgent}
\end{table*}

\begin{table*}[htbp]
    \centering
    \caption{Overall experimental results on APIBank.}
    \resizebox{\textwidth}{!}{
        \begin{tabular}{lccc}
\hline
\multicolumn{1}{c}{\multirow{3}{*}{model name}} & \multicolumn{3}{c}{APIBank}                                                                                     \\
\multicolumn{1}{c}{}                            & \multirow{2}{*}{level-1-given-desc} & \multirow{2}{*}{level-2-toolsearcher} & \multirow{2}{*}{Weighted Average} \\
\multicolumn{1}{c}{}                            &                                     &                                       &                                   \\ \hline
6B- Open-source LLMs                            & \multicolumn{3}{c}{}                                                                                            \\ \hline
Llama-3.2-1B                                    & N/A                                 & N/A                                   & N/A                               \\
DeepSeek-R1-Distill-Qwen-1.5B                   & 9.14                                & 9.20                                  & 9.15                              \\
Qwen2-1.5B-Instruct                             & 47.42                               & 36.41                                 & 45.33                             \\
gemma-2-2b-it                                   & N/A                                 & N/A                                   & N/A                               \\
\hline
\end{tabular}
    }
\label{exp:overall-results-table-APIBank}
\end{table*}

\section{Details of Evaluation Results}
\label{appendix:evaluation-results}
We present all experimental results of the SOTA LLMs on various evaluation metrics across different benchmarks. As shown in Table~\ref{exp:overall-results-table-RoTBench}, Table~\ref{exp:overall-results-table-SealTools}, Table~\ref{exp:overall-results-table-TaskBench}, Table~\ref{exp:overall-results-table-BFCL}, Table~\ref{exp:overall-results-table-Teval}, Table~\ref{exp:overall-results-table-InjectAgent}, Table~\ref{exp:overall-results-table-APIBank}, the overall information provided by the unified benchmarks is insightful and valuable for future development, significantly contributing to our analysis of challenges and opportunities. 

We present one metric per benchmark in Figure~\ref{fig:overall-exp-results} and leave the rest 57 metrics in Appendix~\ref{appendix:evaluation-results}. Specifically, for RoTBench, we use the average Content Filling metric across five datasets, as it measures the model’s ability to fill parameters in tool calls. For SealTools, we report the average F1\_param across three datasets, which reflects parameter-filling capability. For TaskBench, we use the average v\_f1 across three datasets, as it measures filling of tool call parameters (task, parameter name, parameter value). For BFCL, we adopt the overall accuracy as the main metric. For T-Eval, we report the average mean value across all metrics; For InjecAgent, we use ASR-valid (Total), which is in alignment with the original paper.

\begin{figure*}[t]
    \centering
    \includegraphics[width=5.5in]{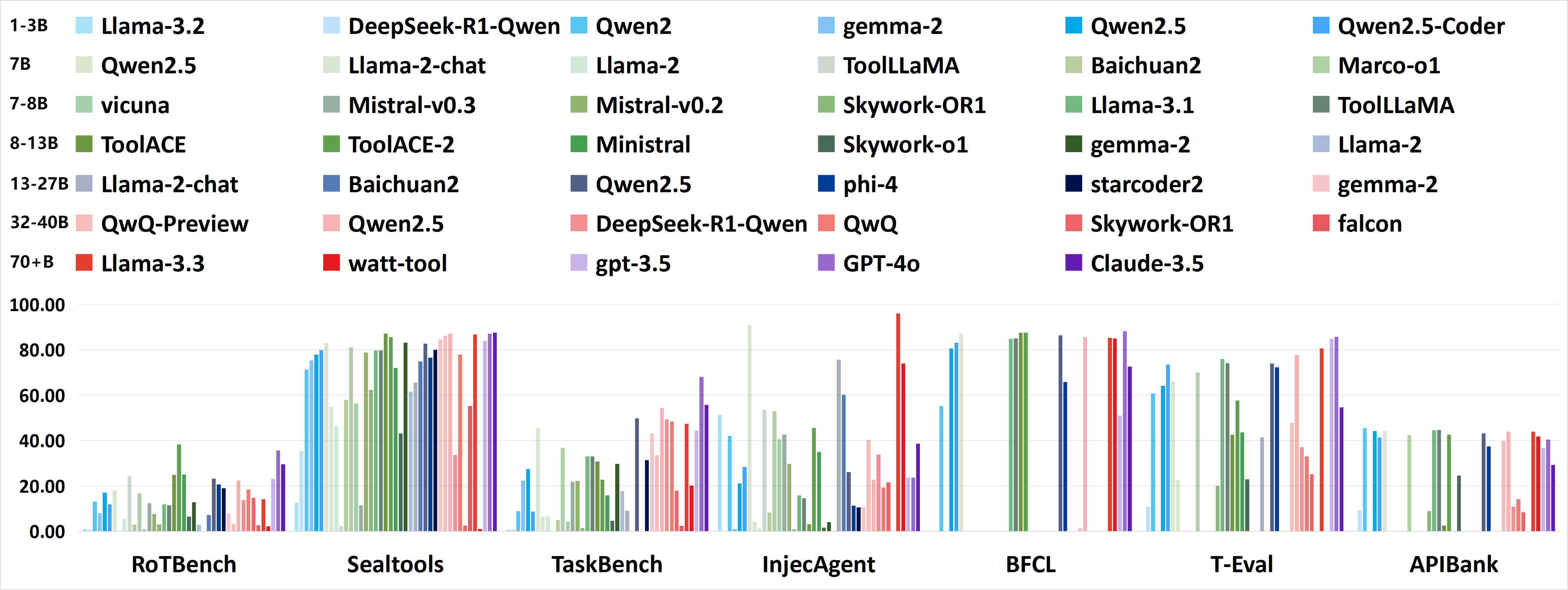} 
    \caption{Overall model performance on our constructed ToLeaP. Due to the limited space, we present one representative evaluation metric per benchmark, covering 7 out of the total 64 metrics. }
    \label{fig:overall-exp-results}
\end{figure*}

Through our constructed one-click-evaluation benchmark set, researchers are able to observe the informative feedback of their developed LLMs as guidance, which pinpoints the limitations and effectively facilitate the future explorations.

\section{Details of Inference Methods}
\label{appendix:inference-methods}
Our inference methods contain two optional inference methods: (1) chat completion, where inputs are transformed into a chat-style format; and (2) streaming method, where inputs are retained in the output and LLM generates text incrementally. We conduct extensive experiments to demonstrate that distinguishing between the two methods is essential: We found that the streaming method is more compatible with the LLaMA-2 model series, while the chat-completion-based method is better suited for LLaMa-3.1 and newer models.

For models that are incompatible with a given benchmark (e.g., models that cannot be evaluated using the original settings of a benchmark), we design and implement inference methods for these models. We categorize all inference methods into chat-completion and streaming methods and select the method with higher performance as the final inference method for each model.

\section{Details of Evaluated Models}
We evaluated both proprietary and open-sourced LLMs at various scales, covering 15 LLMs in total to provide a comprehensive evaluation of their tool-using capabilities. The proprietary LLMs include GPT-3.5-turbo, GPT-4o, and Claude-3.5-Sonnet, while the open-sourced LLMs involve Llama3.1-8B-Instruct, Qwen2.5-7B-Instruct, Mistral-8B-Instruct-2410, ToolACE-8B, Marco-O1, Skywork-O1-Open-Llama-3.1-8B, Llama2-7B-Chat, Llama2-7B, ToolLlama2-7B-v2, Llama2-13B, Llama2-13B-Chat, and QwQ-32B.

\section{Details of Challenge 1}
\label{appendix:challenge-1}
\begin{figure}[htbp]
    \centering
    \includegraphics[width=1\linewidth]{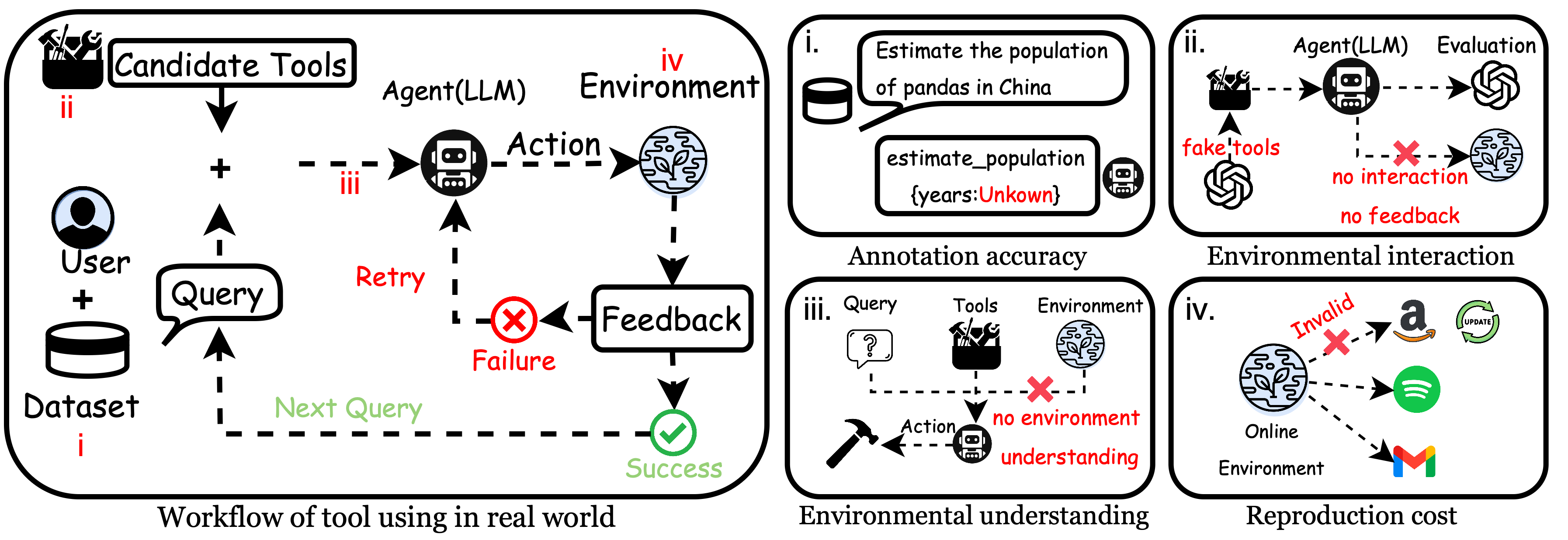} 
    \caption{Drawbacks in the current benchmark in comparison with real-world scenarios. \romannumeral1. Low quality and ambiguous datasets; \romannumeral2. Artificial tools and simulated evaluation; \romannumeral3. limited information for llm; \romannumeral4. Updating online environment and unstable tools.}
    \label{fig:challenge-1-benchmark}
\end{figure}

The typical process of tool using with LLMs in real-world scenarios is illustrated in Figure~\ref{fig:challenge-1-benchmark}. Different benchmarks evaluate different capabilities in the process.

\section{Details of Challenge 2}
\label{appendix:challenge-2}
We compared models trained on our collected full datasets (33 datasets) with those trained only on the publicly available ToolACE training data across four benchmarks. Results in Figure~\ref{fig:toolace-results} revealed a key pattern: although full-data models occasionally achieved strong performance, their training was unstable. To this end, we evaluated both the SFT and DPO methods. On SealTools and RotBench, SFT scores fluctuated dramatically—rising from 58.58\% to 68.47\%, plunging to 62.39\%, then recovering to about 68.42\%. By contrast, DPO reached in just 80 steps the performance that SFT only achieved after 100,000 steps, and converged more quickly and smoothly. Moreover, adding additional data late in training consistently degraded performance, underscoring the need for models to autonomously select what to learn.

\section{Case Study of Challenge 3}
\label{appendix:challenge-3}
\begin{figure}[ht]
    \centering
    \includegraphics[width=0.8\linewidth]{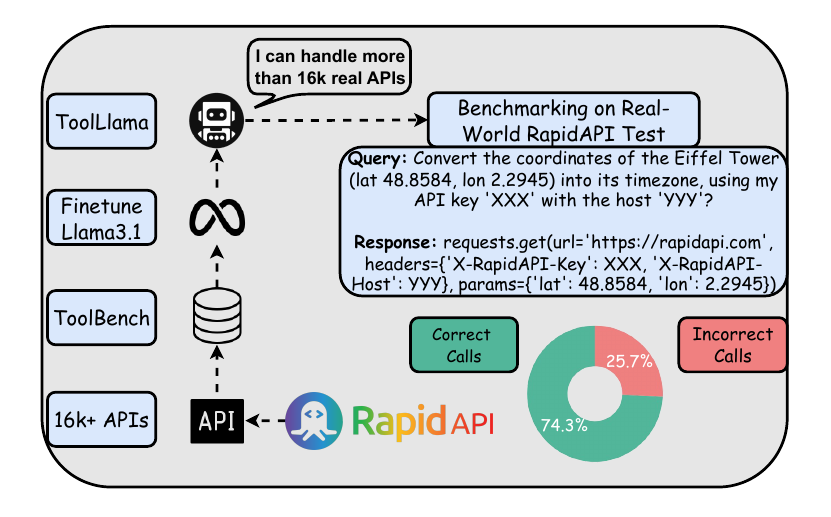} 
    \caption{An example of the generalization ability investigation based on ToolLLaMA.}
    \label{fig:realapi}
\end{figure}

We provide a case study to illustrate the current bottleneck, namely the lack of generalization capability of tool learning with LLMs. As shown in Figure~\ref{fig:realapi}, despite the training process where the LLaMA-3.1-8B-Instruct-based ToolLLaMA successfully master 16000+ APIs from the RapidAPI platform, and the evaluation process where ToolLLaMA demonstrates its expertise in utilizing API tools on ToolBench, it fails to make correct tool-using instructions when given the queries in BFCL with APIs from the same source (RapidAPI). 

\begin{figure}[ht]
    \centering
    \includegraphics[width=0.6\linewidth]{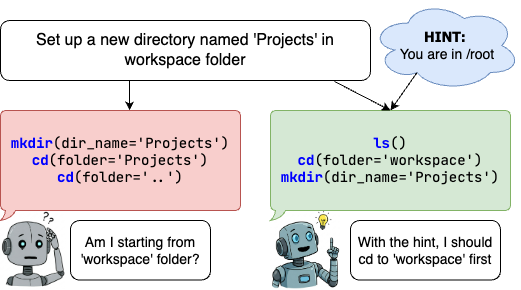}
    \caption{Providing key information before generating answers can improve the tool call generation accuracy for LLMs.}
    \label{fig:bfcl_multi_turn_hint}
\end{figure}

\section{Details of Opportunity 2}

We present the detailed experimental results in Table~\ref{Tab:compatibility}.

\begin{table}[htbp]
\centering
\scriptsize
\setlength{\tabcolsep}{4pt}
\renewcommand{\arraystretch}{1.2}
\caption{Results of compatibility-aware learning of different models on BFCL.}
\resizebox{0.75\textwidth}{!}{
\begin{tabular}{lcccccc}
\toprule
\textbf{Model} & \textbf{Simple} & \textbf{Multiple} & \textbf{Parallel} & \textbf{Para. Multi.} & \textbf{Irrelevance} & \textbf{Average} \\
\midrule
Qwen2.5-1.5B             & \textbf{72.42\%} & \textbf{87.00\%} & \textbf{81.50\%} & \textbf{75.50\%} & \textbf{62.68\%} & \textbf{75.82\%} \\
Qwen2.5-1.5B (Auto)      & 62.92\% & 74.50\% & 52.50\% & 56.00\% & 35.83\% & 56.35\% \\
\midrule
Qwen2.5-3B               & \textbf{73.33\%} & \textbf{92.00\%} & 73.50\% & \textbf{76.50\%} & \textbf{64.26\%} & \textbf{75.92\%} \\
Qwen2.5-3B (Auto)        & 66.83\% & 91.00\% & \textbf{76.50\%} & 72.50\% & 59.58\% & 73.28\% \\
\midrule
Qwen2.5-7B               & 71.83\% & 95.00\% & 90.00\% & 86.00\% & 69.08\% & 82.38\% \\
Qwen2.5-7B (Auto)        & \textbf{76.25\%} & \textbf{96.00\%} & \textbf{90.50\%} & \textbf{89.00\%} & \textbf{73.33\%} & \textbf{85.02\%} \\
\midrule
Llama3.1-8B              & 55.83\% & 54.00\% & 48.50\% & 34.50\% & 4.86\% & 39.54\% \\
Llama3.1-8B (Auto)       & \textbf{77.27\%} & \textbf{84.00\%} & \textbf{73.50\%} & \textbf{74.50\%} & \textbf{10.83\%} & \textbf{64.02\%} \\
\bottomrule
\end{tabular}
\label{Tab:compatibility}
}
\end{table}

\section{Details of Opportunity 3}
\label{appendix:3-opportunity}
\begin{figure}[t]
    \centering
    \includegraphics[width=1\linewidth]{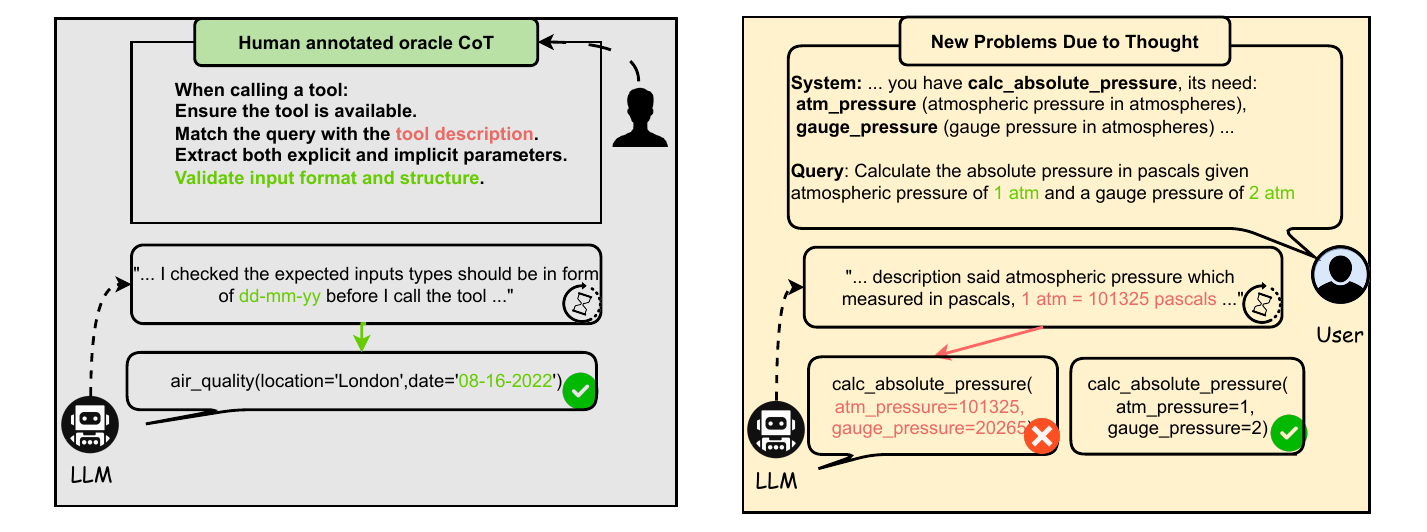} 
    \caption{The performance of LLaMA-3.1-8B-instruct given human annotated tool learning rationale.}
    \label{fig:thought-example}
\end{figure}

We adopt the g1\footnote{\url{https://github.com/bklieger-groq/g1}} method, inspired by OpenAI’s o1 approach that uses Chain-of-Thought-style self-reasoning to enhance model generalization. This method constructs step-by-step reasoning paths to improve the model’s thinking ability. In our experiments, we applied G1 to the BFCL benchmark, prompting the model to generate a sequence of reasoning steps (rather than final answers) for each test case. These "thoughts" were generated using the G1 prompt setting. To evaluate the method more efficiently, we concatenated the generated thoughts back into each test case input, without changing any benchmark configuration. The model then generated final outputs based on the enriched context, combining the original query and the model’s own reasoning. An illustration of opportunity 3 is shown in Figure~\ref{fig:thought-example}

\section{Case Study of Opportunity 4}
We present the case study of the effectiveness of explicit key clues in identifying and recalling. As shown in Figure~\ref{fig:bfcl_multi_turn_hint}, we provide the human-annotated key clues to LLMs without any other explanation and guidance. With the recalling of key clues, LLMs are able to avoid making mistakes and succeed in achieving stronger tool-using performance.

\begin{table*}[htbp]
    \centering
    \caption{First- and second-level benchmarks and reasons for not being merged into our work.}
    \begin{tabularx}{\textwidth}{c >{\raggedright\arraybackslash}p{0.25\textwidth} X}
\hline
& Benchmark & Status \\ \hline
1-7 & Taskbench, T-Eval, SealTools, RoTBench, InjecAgent, BFCL & Merged. \\ \hline
8 & API-Bank & The original paper only implemented evaluation methods for closed-source models and did not provide interfaces for open-source models. We modified it using the offline vLLM interface. The code repository mentions a package conflict with googletrans, so related tasks did not run, which led to some deviation in the scores. \\
9 & ToolBench1 & This codebase is built on real APIs, but due to its age, it has not been updated to match API version iterations. As a result, it’s difficult to integrate it into current benchmarks. \\
10 & ToolAlpaca & This codebase is also based on real APIs, but it does not fully provide the required parameters, and some parameter acquisition methods are not clearly defined, making it hard to reproduce results. \\
11 & ToolEyes & Most metrics were evaluated using GPT-4o’s simulated assessments and tested the ability to call pre-written Python scripts. However, due to severe conflicts with our environment (which differs significantly from the original one) and changes in closed-source models, our results couldn't align with the original. We decided not to include this evaluation in the final assessment. \\
12 & ToolQA & This codebase is built on real APIs, but the source data is highly fragmented and vast in volume. The tools involve complex environment interactions, and the provided code lacks maintenance, making it hard to run in our current environment. \\
13 & ToolEmu & The code uses vLLM to load open-source models, but the environment setup files do not include vLLM. Installing vLLM caused conflicts with our existing environment, preventing the code from running properly. We replaced vLLM with HuggingFace’s pipeline. As closed-source models were inconvenient to use, we changed the interface. However, the loading process for both models did not align well with the subsequent inference methods, requiring extensive modifications. So, we ultimately chose not to include this evaluation in the final benchmark. \\
14 & StableToolBench & The original code version is outdated, complex, slow in inference, and inaccurate in output, making it hard to extract key information. Switching to vLLM inference failed due to large version gaps. Switching to HuggingFace pipeline further slowed down an already slow process. Hence, we chose not to include it in our final evaluation. \\
15 & ToolLens & Did not propose a new benchmark but relies on the ToolBench dataset. \\
16 & ShortcutsBench & Based on this repo, it interacts with APIs from the Apple App Store, requiring the appropriate OS, Apple account, and testing platform. We found that due to firewall and system limitations, we cannot support real dataset interaction. \\
17 & ToolSandbox & Uses RapidAPI and a special data format (milestone) for evaluation. When we tried to adapt it for vLLM-based testing, version conflicts with other benchmarks occurred, preventing further progress. \\ \hline
\end{tabularx}
\label{exp:merge-table-tier12}
\end{table*}

\begin{table*}[htbp]
    \centering
    \caption{Third-level benchmarks and reasons for not being merged into our work.}
    \begin{tabularx}{\textwidth}{c >{\raggedright\arraybackslash}p{0.25\textwidth} X}
\hline
& Benchmark & Status \\ \hline
18 & APIBench & We found no way to locate or run an evaluation program to assess model performance on its tasks. APIBench is Gorilla's own training dataset and lacks a defined inference method and evaluation code. \\
19 & RestBench & As the authors did not provide evaluation code or ground-truth data, we were unable to reproduce the results. After submitting an issue, we did not receive any response that could resolve the problem.\\
20 & ToolBench2 & We successfully reproduced the code up to the inference output stage, but due to a lack of maintenance, the APIs used were outdated, preventing smooth progress. \\
21 & MetaTool & Similar to RestBench, the authors did not provide evaluation code or ground-truth data. We couldn’t reproduce the results and received no helpful response after raising an issue. \\
22 & NexusRaven & We attempted to reproduce the code, but the links provided in the repository were inaccessible. After submitting an issue, we did not receive any response that resolved the problem. \\
23 & UltraTool & We couldn’t find or run any evaluation program to measure the model’s performance. UltraTool seems more like a training dataset rather than a benchmark. \\
24 & API-BLEND & This project collected and reformatted a large number of old tool-call datasets. However, the original repository only provides data conversion scripts and does not include any evaluation code. Therefore, we were unable to reproduce the original conclusions. Critical information is missing, preventing effective evaluation and validation. \\
25 & ToolTalk & This project is relatively old. While trying to reproduce the code, we encountered a ConnectionError, likely due to outdated code, making it hard to integrate into the current benchmark. \\
26 & ToolSword & This project only open-sourced the data but not any code, so we couldn't reproduce the results. \\
27 & SciToolBench & This project did not open-source critical code or datasets. \\
28 & VIoT & During reproduction, we encountered the error \texttt{AttributeError: DeepSpeedCPUAdam object has no attribute ds\_opt\_adam} and couldn’t find a good solution. \\
29 & MLLM-Tool & This is a multimodal benchmark. We may consider merging it in the future. \\
30 & GeoLLM-QA & During reproduction, we encountered the error: Error while generating prompt 8: HTTPSConnectionPool...Network is unreachable, and couldn’t resolve it. \\
31 & SoAyBench & Files mentioned in the repository README, such as \texttt{api\_graph.py}, do not exist, and \texttt{functions.py} has critical errors that prevent it from running. \\
32 & GTA & Multimodal; This is a multimodal benchmark. We may consider merging it in the future. \\
33 & WTU-Eval & No GitHub repository was provided. \\ \hline
\end{tabularx}

\label{exp:merge-table-tier3}
\end{table*}

\section{Details of the Reproduction Notes}
We explain the exact reasons why the 26 benchmarks are not involved in the Tier I category, failing to be merged in our constructed one-click evaluation framework ToLeaP. The reasons are shown in Table~\ref{exp:merge-table-tier12} and Table~\ref{exp:merge-table-tier3}.

\end{document}